\pdfoutput=1

\documentclass[11pt]{article}

\usepackage{EMNLP2023}

\usepackage{times}
\usepackage{latexsym}

\usepackage[T1]{fontenc}

\usepackage[utf8]{inputenc}

\usepackage{microtype}

\usepackage{inconsolata}
\usepackage{graphicx}
\usepackage{algorithmic}
\graphicspath{ {./images/} }
\usepackage{multirow}
\usepackage{algorithm}
\usepackage{booktabs}       

\title{Hallucinations in Neural Automatic Speech Recognition: Identifying Errors and Hallucinatory Models}

\author{Rita Frieske \\
  HKUST \\
  \texttt{rmfrieske@connect.ust.hk} \\\And
  Bertram E. Shi \\
  HKUST \\
  \texttt{eebert@connect.ust.hk} \\}

\begin{document}
\maketitle
\begin{abstract}
Hallucinations are a type of output error produced by deep neural networks. While this has been studied in natural language processing, they have not been researched previously in automatic speech recognition. Here, we define hallucinations in ASR as transcriptions generated by a model that are semantically unrelated to the source utterance, yet still fluent and coherent. The similarity of hallucinations to probable natural language outputs of the model creates a danger of deception and impacts the credibility of the system. We show that commonly used metrics, such as word error rates, cannot differentiate between hallucinatory and non-hallucinatory models. To address this, we propose a perturbation-based method for assessing the susceptibility of an automatic speech recognition (ASR) model to hallucination at test time, which does not require access to the training dataset. We demonstrate that this method helps to distinguish between hallucinatory and non-hallucinatory models that have similar baseline word error rates. We further explore the relationship between the types of ASR errors and the types of dataset noise to determine what types of noise are most likely to create hallucinatory outputs. We devise a framework for identifying hallucinations by analysing their semantic connection with the ground truth and their fluency. Finally, we discover how to induce hallucinations with a random noise injection to the utterance.

\end{abstract}

\section{Introduction}

Recent development of neural architectures brought the improvement of performance among many natural language processing (NLP) tasks, including in automatic speech recognition (ASR). Deep learning architectures that require large amounts of data to achieve optimal performance come, however, with challenges specific to their kind that are not observed in parametric models~\cite{koehn-knowles-2017-six}. One such challenge is the appearance of hallucinations, that we define as fluent and coherent outputs of neural models entirely disconnected from the input. 

The phenomenon of hallucinations can be observed among different NLP tasks such as machine translation, dialogue systems, question answering, summarization or captioning. Nowadays, with the public deployment of Large Language Models (LLMs) such as ChatGPT, the topic of hallucinations has gained vast attention beyond the scientific community~\citep{Alkaissi2023ArtificialHI}. The lack of control over the data that are used to train LLMs increases the probability of creation of hallucinatory outputs. Furthermore, the blurry definition of hallucinations and lack of quantitative theoretical frameworks that would help to distinguish them from other errors impacts the research on finding methods of their removal. Hallucinations are a dangerous model error because of their lack of connection with the model input, but fluent and coherent manner that makes them look like a probable model output. This may confuse and deceive users and creates lack of trust to the model and, in case of publicly deployed LLMs - to the industry. Majority of the scientific articles on hallucinations cover various NLP tasks. There is, however, to our knowledge no literature coverage on this problem for task such as ASR or spoken language understanding (SLU). We try to cover this research gap by addressing specific ways that hallucinations in ASR can be induced through addition of random noise to the utterance, and explore the ability to detect hallucinations through the implementation of an algorithm that produces hallucinations during the testing phase that helps to determine  susceptibility of the model to hallucinations.

\begin{table}[h]

\resizebox{\columnwidth}{!}{%
\centering
\begin{tabular}{@{}lllll@{}}
\toprule
Libri 360    & dev-clean & dev-other & test-clean & test-other \\ \midrule
BASELINE WER & 6.14      & 16.44     & 7.27       & 15.65      \\
RR WER       & 8.56      & 21.02     & 10.71      & 21.34      \\
RU WER       & 6.5       & 17.69     & 6.99       & 17.61      \\
UU WER       & 7.25      & 20        & 8.47       & 19.8       \\
UR WER       & 14.83     & 39.32     & 16.78      & 38.69      \\ \bottomrule

\end{tabular}%
}
\caption{Word error rate (WER) of models trained on LibriSpeech 360: baseline and the models with mismatched labels: repeat-repeat (RR), repeat-unique (RU), unique-unique (UU) and unique-repeat (UR). The scores show small differences apart from UR model, which often copies repeated sequences from the training dataset as a test output.}
\label{tab:wer_noisy_data}
\end{table}
Errors in ASR are an important measure of the system performance. The main difference between phonetic ASR errors and hallucinations lies in the severity of the latter. Phonetic ASR errors appear as badly transcribed words or phrases, especially when utterances are phonetically similar. They are being evaluated as a number of phonetic substitutions, insertions and deletions. On the contrary, the hallucinatory output does not have phonetic or semantic connection with the source utterance, even though it is often fluent and coherent. The latter feature makes it especially dangerous in commercially deployed models, since the end-users may not be monitoring the source utterance, e.g. during transcription of lectures or speeches (see: table \ref{tab:definition} to compare ASR errors).

\begin{table*}[t]
\centering

\label{tab:word_error_hallucination}
\resizebox{\textwidth}{!}{%
\begin{tabular}{@{}llll@{}}
\toprule
 & Phonetic Error & Hallucination & Oscillation\\ \midrule
Reference & \begin{tabular}[c]{@{}l@{}} millimeter roughly \\ one twenty fifth of an inch  \end{tabular} & \begin{tabular}[c]{@{}l@{}}captain lake did not look at all \\ like a london dandy now     \\     \end{tabular}  & indeed ah\\

Output & \begin{tabular}[c]{@{}l@{}}miller made her roughly \\ one twenty fifths of an inch \end{tabular} & \begin{tabular}[c]{@{}l@{}} \\ will you let annabel ask her if she\\  sees what it is you hold in your\\  arms again the voice was soft  and wheedling \\ no annabel said  aunt rachel faintly \\ will you  rock again\end{tabular} & \begin{tabular}[c]{@{}l@{}}  a ay indeed ay ay ay ay \\ ay  ay ay ay ay ay\end{tabular} \\ \bottomrule
\end{tabular}
}
\caption{Examples of differences between the  common ASR errors and hallucinatory content}
\label{tab:definition}
\end{table*}

Similarly to BLEU score, a metric for machine translation (MT) tasks, typical automatic metrics of an ASR system such as word error rate (WER) do not reflect the difference between phonetic errors and hallucinations \cite{Lee2018HallucinationsIN} (see: table \ref{tab:wer_noisy_data}). It is therefore important to find the way of capturing such errors and evaluating models that have increased tendency to generate them.

We propose a framework to differentiate hallucinations from other ASR errors based on how semantically related to the reference transcription they are. Similarly, we assess the fluency of the output sequences to determine whether the outputs are probable model outputs, or just completely erroneous transcriptions. Since hallucinations should resemble probable model outputs, the fluency measure should be high and semantic connection to reference should be low.
 
The main contributions of this work are as follows:
\begin{itemize}
    \item we present the algorithm that assesses whether the model is susceptible to hallucinations. Our method does not require access to the training dataset
    \item we implement a framework to identify hallucinations and differentiate them from ASR phonetic errors, showing a clear distinction between the two error types
    \item we present the method to induce hallucinations with random noise injection to the source utterance
    \item we show correlation between ASR dataset mismatch and ASR error distributions. We explore how much label noise is needed to produce given error distributions 
    \item we show that WER is not sensitive towards different ASR error types 

\end{itemize}
Furthermore, to our knowledge, this is the first work on hallucinations understood as a semantically disconnected and fluent output error of an ASR system.
 Understanding hallucinations is crucial to model security in ASR and NLP tasks.

\section{Related work}
\label{sec: related_work}

There is no one clear definition of hallucinations in a current body of work, \cite{JiZiwei} distil the definition that spreads across several natural language generation (NLG) tasks such as dialogue, question answering or machine translation. They describe hallucinations as "the generated content that is
nonsensical or unfaithful to the provided source content". They further divide hallucinations into intrinsic and extrinsic. Intrinsic hallucinations contradict the source content, and extrinsic cannot be verified from the source content.

The dangerous feature of hallucinations in neural models is their fluency and coherency, which makes them dangerous in fact-checking, knowledge grounded dialogue or machine translation \cite{koehn-knowles-2017-six}.

As per our knowledge, to date, there are no articles on hallucinations in neural ASR models, given that hallucinations follow the above-mentioned definition. \citet{Serai} use the term "hallucination" in their work in the context of ASR error creation, rather than referring to the specific type of error. Similarly, \cite{Sagae} use the term "hallucinate" to refer to producing ASR errors.
Concurrently to the literature of hallucinations in ASR, phonetic ASR errors (in literature referred to just as errors) are well researched. \cite{Anguita} analysed likely confused phone distances in HMM models, \cite{Jyothi2010DiscriminativeLM} built a phoneme confusion matrix to determine how often input text can be confused. \cite{shivakumar_li_knight_georgiou_2019} present multiple example of phonetic errors in ASR as well as the methods of mitigation, such as lattice rescoring or minimum error rate training (MERT). \cite{fang-etal-2022-non} propose method of ASR post-processing error correction by using non-auto-regressive phonologically based approach.

Evaluation of hallucinatory content is a challenge across different NLP and ASR tasks. To detect fluent but nonsensical outputs \cite{martindale-etal-2019-identifying} propose BVSS Metric, which is based on cosine similarity and measuring proportion of information in output that is also in the reference. \cite{ERRATTAHI201832} lists metrics alternative to WER in ASR such as RIL and its approximation WIL based on information theory and measure statistical dependence between the input and output. Regarding fluency measures that could be helpful to assess hallucinatory output, the proposed methods are mostly catering to other NLP tasks such as ROUGE metric for summarization \cite{lin-2004-rouge}. Although ROUGE is currently the most popular metric, it requires the access to the parallel corpus. Alternatively, \cite{mutton-etal-2007-gleu} propose a metric for sentence level fluency evaluation called GLEU.


The algorithm implemented in this article is based on the one implemented in the machine translation context by \citet{Lee2018HallucinationsIN}. 
Regarding the experimental noisy models, the original recipe was provided by~\citep{Raunak2021} for MT. We adjusted the noise amount for the needs of ASR experiment. The concept of error stimulation with injecting noise to dataset is not new and in the context of ASR proposed by \cite{Fetic2021TopicMR} who also implement similarity metric to evaluate their findings. \cite{morales2007adding} add noise to improve the robustness of the ASR models. Finally, \cite{Iwamoto2022HowBA} show the influence of acoustic artifacts on produced errors by performing error decomposition analysis.


\section{Methodology}
To explore the hallucinations in ASR, we create models that are prone to hallucination generation. We do that by creating noisy datasets with mismatched labels. To evaluate hallucinatory susceptibility of the models, we compare hallucination distributions before and after random noise injection to the source utterances of an evaluation set. In order to identify hallucinatory output, we evaluate its semantic similarity to the reference and its fluency.

\subsection{Datasets}

\begin{table}[b]
\centering

\resizebox{\columnwidth}{!}{%
\begin{tabular}{lrr}
\hline
\textbf{Dataset Split} & \multicolumn{1}{l}{\textbf{Utterances}} & \multicolumn{1}{l}{\textbf{Hours}} \\ \hline
LibriSpeech Train 360        & 104014 & 363.6    \\
8\% Label Noise  & 10000  & $\sim$34 \\
16\% Label Noise  & 20000  & $\sim$68 \\
\hline
\end{tabular}%

}
\caption{Statistics of the LibriSpeech 360 used to train the models, and of noisy splits that were added to the main training set. For the experiments analysing pattern distribution of ASR errors (see: \ref{sec:ASR_error_distribution}) the noise subsets were added to the main training set. }
\label{tab:dataset_stats}
\end{table}
The dataset used for the most of the experiments in this work is a widely used LibriSpeech dataset \cite{librispeech}, that consists of utterances extracted from the audiobooks. Since processing of speech is computationally expensive, we use a smaller split of 360 hours of speech out of 960 hour dataset. For evaluation, we use test and development splits divided into clean speech, in which the lector's voice is clear and understandable, as well as splits marked as 'other', that are generally more difficult to understand. The reason behind choosing LibriSpeech is that it is acted speech recorded in silent environment, therefore it is easy to observe degradation of the result upon addition of the noise.

\begin{table}[h]
\centering

\resizebox{\columnwidth}{!}{%
\begin{tabular}{lr}
\hline
\textbf{Dataset Split} & \multicolumn{1}{l}{\textbf{Hours}} \\ \hline
CommonVoice 4        & 1,119    \\
CommonVoice DeltaSegment 10.0,12.0,13.0  & 161 \\
\hline
\end{tabular}%

}
\caption{Statistics of the CommonVoice Dataset}

\label{tab:covost_stats}
\end{table}
For generalization, we also use a model trained on Common Voice Corpus 4, English language split, that consists of 1119 hours of validated read speech from the Wikipedia entries \cite{ardila-etal-2020-common}. For dataset noise through label mismatch and for evaluation we use Common Voice Delta Segment 10.0,12.0 and 13.0. Since Common Voice dataset is a crowdsource project, the utterance quality varies from professionally recorded LibriSpeech, which is reflected in the overall higher WER.

\subsection{Dataset Noise through Label Mismatch}\label{noise}

We explore correlation between ASR error distribution and datasets with partially noisy mismatched labels. ASR models require huge datasets for training. Creating a large, balanced and supervised audio dataset is both expensive, and in cases of low resource languages, very difficult. For crowdsourced datasets such as CommonVoice, only parts of the dataset are validated by the users. Lack of audio data can be mitigated with data augmentation \cite{Park_2019}, creating synthetic audio data via voice conversion or text-to-speech (TTS) \cite{thai2019}, and by using one ASR model to create a training set for another one \cite{Ma2023}. Such methods of increasing data volume might create dataset imbalance that impacts generation of certain types of errors as a result. 

To discover how different types of label mismatch impact error types in ASR, we follow the dataset mismatch recipe described by \citet{Raunak2021} for the MT task and create four noisy subsets. The volume of each subset is approximately 8\% or 16\% of the original dataset, depending on the experiment. The type of noise added to the dataset is also a name of each model trained on it:

\begin{itemize}
    \item Unique-Unique (UU): random unique source sentences paired  with an unrelated unique random target sentences
    \item Repeat-Repeat (RR): unique source sentences paired with unrelated unique random target sentences. 10 such pairs are then repeated, to reach the desired number of noisy utterances
    \item Repeat-Unique (RU): repeating source utterances paired with unrelated unique random target sentences 
    \item Unique-Repeat (UR): unrelated unique random source utterances paired with repeating targets
\end{itemize}

The noisy subsets are then added to LibriSpeech 360 dataset. 
Afterwards, we use the same transformer model with the same parameters to train a baseline and four separate noisy models denoted by the above abbreviations. 



\subsection{Evaluation}

\begin{figure*}[hbt!]
  \centering
  \includegraphics[width=1\textwidth]{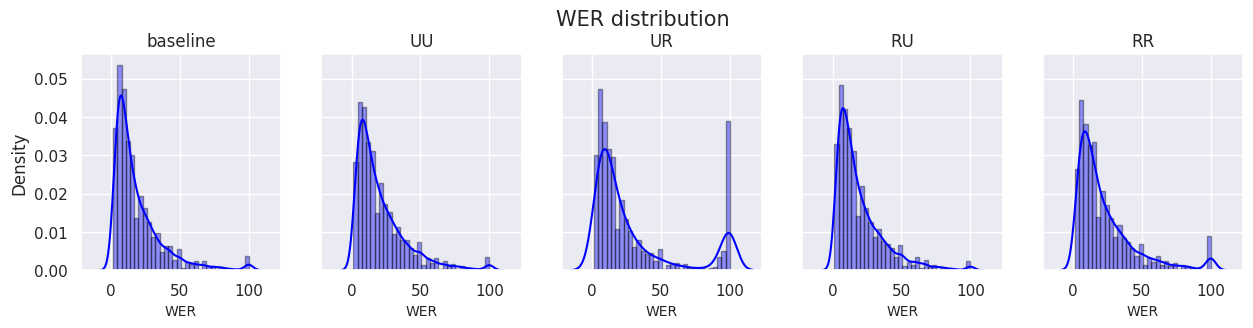}
  \caption{Word Error Rate (WER) distribution across evaluation dataset of assessed models. We exclude 0 WER scores since they are not considered error and for histogram clarity. Although the WER scores alone are not informative with regard to the types of errors produced by the models, the WER distribution in UR and RR model shows increase of highly erroneous outputs.}
\end{figure*}

\paragraph{ASR Model Evaluation}
Hallucinations are not reflected in certain NLP metrics, such as BLEU scores for machine translation task~\citet{Lee2018HallucinationsIN}. To confirm whether hallucinations have an impact on WER, we compare it across hallucinatory and non-hallucinatory models. Word error rate is formulated as:
 \begin{equation} WER=\frac{S+I+D}{N} \times 100\% \end{equation}
Where S indicates a number of substitutions, I stands for insertions, D for deletions and N is a vocabulary count. Since WER is a distance based ASR metric, we also compare BLUE and CHRF2 scores as examples of overlapping n-gram based metrics to validate whether any of them is more sensitive to certain error types than the other. 
Finally, we compare a distribution of a sentence scores across evaluation datasets for a more fine-grained picture of the error distribution.

\paragraph{Semantic Relationship} We use cosine similarity to estimate if the reference and the output are semantically related. This metric is chosen because its ability to find sequence vectors that are closest to each other in the Euclidean space without taking into account the length of each sequence. If o is an output vector and r is a reference vector, then:
 \begin{equation} \cos ({\bf o},{\bf r})= {{\bf o} {\bf r} \over \|{\bf o}\| \|{\bf r}\|} = \frac{ \sum_{i=1}^{n}{{\bf o}_i{\bf r}_i} }{ \sqrt{\sum_{i=1}^{n}{({\bf o}_i)^2}} \sqrt{\sum_{i=1}^{n}{({\bf r}_i)^2}} } \end{equation}

The sensitivity to the semantics of the sentence and agnosticism towards its length, makes cosine similarity a good metric to distinguish hallucinations from oscillations and phonetic errors. We define hallucinations as semantically disconnected outputs. Therefore, they are the errors with the lowest cosine similarity and simultaneously high WER. 

 We characterise ASR phonetic errors by incorrect transcription of particular sounds that are phonetically similar to the reference word, or are misinterpreted because of wrong clustering of the spoken utterance. Such errors return moderately high or high cosine similarity scores, along with a high WER, and the semantically important part of the sentence remains intact. To improve detection of phonetic errors, we explored syllable tokenization of output and reference sentence for more fine-grained cosine similarity score calculation, but resigned from this approach, since the method falsely assigns high cosine similarity to long sentences regardless of their semantic connection. 
 
 Oscillations are the errors that often contain the large part or even the whole correctly transcribed utterance with an addition of a repeating n-gram, that might extend the length of transcription compared to a reference, which results in a high WER. Oscillations return high cosine similarity, showing semantic relationship between the output and the transcription.

\paragraph{Fluency } Hallucinations are erroneous, semantically disconnected and fluent outputs.  This third aspect is crucial to differentiate hallucinations from random repetitions and word salad. To evaluate sentence fluency, we use perplexity, which gives intuition of how viable is the sentence according to LLM. If P(W) is a probability of a sentence given language model, then we calculate perplexity as:
\begin{equation}
    PPL(W)= \Big(\frac{1}{P(W)}\Big)^{(\frac{1}{n})}
\end{equation}
Where n is a number of words in a sentence. Compared with n-gram based fluency metrics, the advantage of this method is taking the whole context of a sentence into account. The disadvantage of perplexity as a metric is its lack of an upper bound, hence we experience the number of outlying values when comparing large number of data. We compare sentence perplexity scores from two LLMs: GPT2 and Google's Flan T5 Small. We find that results returned by Flan T5 Small contain less extreme values of perplexity values than GPT2, hence the decision to choose the former model for the fluency evaluation~\citet{flan5}. 

We also explore ROUGE score as another metric for fluency evaluation. This state-of-the-art summarization metric is often mentioned as the best sentence fluency indicator~\citet{lin-2004-rouge}. 

We find fluency metrics less useful in error identification than cosine similarity, and hence, we use it as an auxiliary metric to exclude sentences that are not fluent rather than to define errors.

\paragraph{Origins of Hallucinations}
Finally, to evaluate the correlation of hallucinations given the training data, we use distance metrics between hallucinated vector and TF-IDF representation of the training data. This way, we detect the 5 closest candidates to each hallucinated sequence. This step is a sanity check to determine whether the hallucinated sentence or its part is simply copied from the training dataset due to label repetitions, or it was independently generated by the model.

\subsection{Induction of Hallucinations}
The volume and the duration of added noise play an important role in creation of an error distribution that reflects the noise added to the training subset and therefore allows for the assessment of the model.
We propose the following perturbations applied to the speech:
\begin{itemize}
    \item random noise injection at the beginning of the sentence. We follow findings of \cite{Lee2018HallucinationsIN}, who determined that input of a token in the beginning of a written sentence causes the most cases of hallucinations
    \item  random noise injection into the whole utterance. This method was previously explored in works on phonetic error generation \cite{Iwamoto2022HowBA} 
\end{itemize}

We follow the theory of \cite{Feldman2020} explored by \cite{Raunak2021} who used perturbations to detect models prone to memorisation that were most likely to produce hallucinations.  We hypothesise that the injected noise should be strong enough, to deteriorate only a number of overly memorised sentences, but not as strong as to deteriorate all of them. In both noise injection scenarios, we explore two digital amplitudes of added noise (0.1 and 0.5 on a scale $[-1,1]$). Furthermore, for noise injection at the beginning of the sentence, we explore different time increments of noise (0.5s and 1s). 

\subsection{Model Evaluation}

We devise a method of finding hallucinations in test time to assess the hallucinatory susceptibility of the model. To achieve that, we use an algorithm that is an extension of the idea of \citet{Lee2018HallucinationsIN} used by them in the MT task. The algorithm uses perturbations understood as alterations to the source utterances to assess the model's potential to produce hallucinatory content. Specifically, we inject random noise at the beginning of each utterance to provoke errors. The idea to perturb an utterance at the beginning comes from the research of \citet{Lee2018HallucinationsIN}, but there, it meant randomly sampled words in a textual input and not a signal alteration. 

\begin{algorithm}[hbt!]
\caption{Hallucination Detection Algorithm}\label{alg:hall_detect}
\begin{algorithmic}
\REQUIRE{Evaluation Data X}
\ENSURE{Hallucinations}
 \STATE  $x\in X$ $y\in Y$;
 \STATE  $threshold\ {WER} \ : t_{wer}$;
 \STATE  $threshold \ cosine \ similarity \ : t_{cos}$;
  \STATE  $threshold\ perplexity\ : t_{ppl}$;
 \WHILE{ Eval  $\hat{y}=model(x)$}
  \STATE {WER} score= {WER}($\hat{y})$;
  \IF{{WER} score$ < t_{wer}$}
  \STATE $\hat{y'}=perturb(x) $;
  \STATE {WER} score'= {WER}($\hat{y'})$;   
                \IF{ {WER} score'$> t_{wer}$}
                \STATE $cosine \ similarity=\cos({\bf\hat{y'}},{\bf y})$;
                \STATE $perplexity=perplexity({\bf\hat{y'}})$;
                \IF{ cosine similarity$ < t_{cos} $ and perplexity $ < t_{ppl}$}
                \STATE hallucinations from perturbation ; 
                \ENDIF
                \ENDIF
 \ELSE{}   
                \IF{ {WER} score$> t_{wer}$}
                \STATE $cosine \ similarity=\cos({\bf\hat{y}},{\bf y})$;
                \STATE $perplexity=perplexity({\bf\hat{y}})$;
                \IF{ cosine similarity$ < t_{cos} $ and perplexity $ < t_{ppl}$}
                \STATE natural hallucinations  ; 
                \ENDIF
                \ENDIF
  \ENDIF
 \ENDWHILE
\end{algorithmic}
\end{algorithm}
The decision on which utterances should be perturbed is based on a scoring of the transcribed result, where the low scoring utterances are filtered to avoid perturbing originally erroneous sequences. After applying perturbations, only the lowest scoring sequences are evaluated for semantic relationship and for fluency. To evaluate hallucinatory susceptibility of a model, we compare the number of filtered hallucinations before and after perturbation and normalize it with the total vocabulary. We discover that hallucinatory models return lower amounts of hallucinatory outputs before than after perturbation. Therefore, the hallucinatory model output returns negative results if perturbation hallucination scores are deducted from non-perturbed hallucination scores. 
This hallucination detection framework, as well as noise injection and error distribution scoring, are all an extension of the original algorithm.

\section{Experiments and Results}
\subsection{Model and hyperparameters}
The model used for all the experiments is an off-the-shelf transformer s provided by fairseq~\cite{ott2019fairseq}. For model architecture details and hyperparameters, see: Appendix \ref{sec:appendix_A1}. As a language model for calculating perplexity, we use FLAN T5 Small. We use unigram ROUGE for calculating ROUGE scores.


\subsection{ASR Error Distribution}\label{sec:ASR_error_distribution}

\begin{figure}[]
  \includegraphics[width=1\columnwidth]{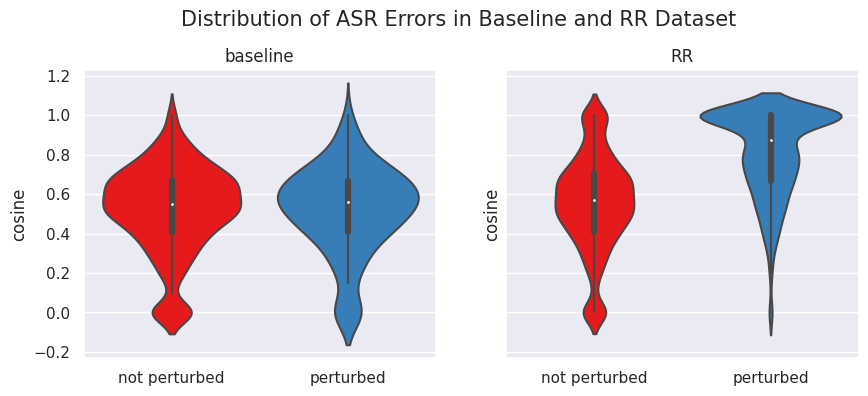}
  \caption{Comparison between cosine similarity distributions before and after perturbations in baseline and RR model. The higher mean of cosine similarity metric in RR model after perturbation suggests an increase of the oscillations, due to the noise from repetitive labels.}
  \label{fig:RR_oscillation}
\end{figure}

We trained 4 models with 8\% of the training set mismatched according to the recipe in section \ref{noise}. The WER, BLUE and CHRF2 scores among different models were similar, or just slightly worse than the baseline apart from the unique-repeat (UR) model, that has shown a clear drop in performance (see: Appendix \ref{sec:appendix_A3}. After error analysis, we discovered that most of the generated errors in UR model were the repeating sequences from the noisy subset. We discovered that the RR model is the most likely to generate oscillations (see: Figure \ref{fig:RR_oscillation}), which correlates with findings of \cite{Raunak2021}. Finally, we observed the increase of hallucinations from perturbations vs without perturbations in UU model in all the noise injection scenarios. The number of hallucinations per dataset given 8\% of injected noise was small (between 1-2\% of all evaluation datasets, apart from UR in which hallucinations accounted for 8\% of the whole dataset). Given the small amount of hallucinations, we increased the amount of noise in the dataset to 16\% which resulted in clear error patterns.

\begin{figure}[]
  \centering
  \includegraphics[width=1\columnwidth]{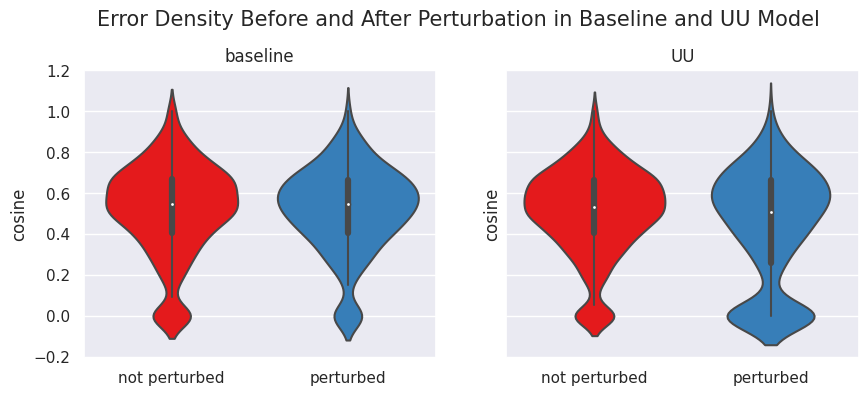}
  \caption{The error distribution in baseline and UU model before and after running the hallucination detection algorithm. The lower peak of the distribution near 0 cosine similarity indicates hallucinatory outputs. The increase in hallucination ration in Unique-Unique dataset is not reflected in its WER results. }
  \label{fig: UU_detection}
\end{figure}

 We observe that the increased amount of repetitive labels in the dataset correlated with oscillations. We discover that large amounts of repeating target sequences paired with unique sequences cause hallucinations that are copies from the repeated training labels. We find that the uniquely mismatched labels both for source and target sequences (UU model) increase hallucinatory susceptibility of the model. This susceptibility is not reflected in the model's overall score, but the probability of hallucination increase is higher than in other models when encountered with noisy samples (see: Figure \ref{fig: UU_detection}).
 
\subsection{Hallucination Identification}\label{sec:identification}
We discover that the cosine similarity is sensitive towards error distribution. In most of the analysed cases, we observe clear distinction between hallucinatory outputs and phonetic errors. After analysing error distribution peaks, we set the threshold between the hallucinatory output and phonetic errors as 0.2 cosine similarity. 
\begin{figure*}[h]
  \centering
  \includegraphics[width=1\textwidth]{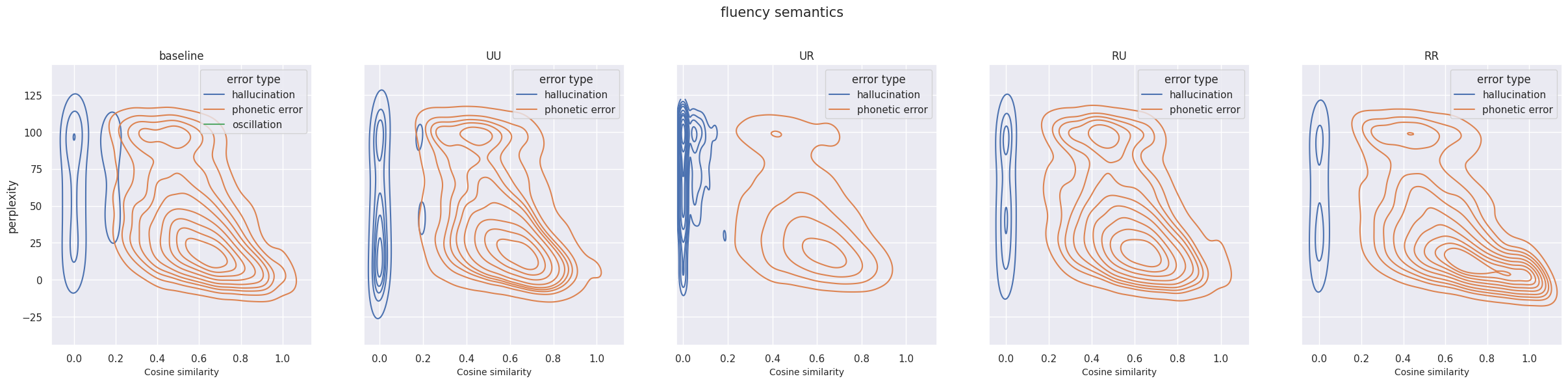}
  \caption{The cosine similarity and perplexity kernel distribution across 5 models, showing clear distinction between phonetic errors and hallucinations. }
\end{figure*}
We use similar heuristic method to define threshold for WER as 30 and perplexity as 200 to evaluate fluency of a sentence.
The perplexity and ROUGE as fluency metrics do not indicate error distribution but rather help exclude outputs that are not fluent and nonsensical such as word salad, repeating syllables etc. 

\subsection{Hallucination Induction}\label{sec:induction}

Dataset label mismatch can increase the number of errors across datasets, creating general hallucination distribution patterns. To increase probability of creation of hallucinatory output from the model with small or medium amount of label noise, which is the most possible real life scenario, the test utterance should be noisy (see: Appendix \ref{sec:appendix_A2})
We discover that the noise injection at the beginning of the utterance has the influence on hallucination distribution, while the noise injected throughout the whole utterance does not. The noise injected equally throughout the sentence impacts however performance of the whole evaluation dataset, and it is reflected in the distribution of phonetic errors (see: Appendix \ref{sec:appendix_A4}). We also note that the length and volume of the random noise injected at the beginning of the sentence has slight influence on the hallucination production, we chose 1s long random noise perturbation of amplitude 0.5 as the most successful one with LibriSpeech dataset. 

\paragraph{Evaluation of the models with Hallucination Detection Algorithm} 
 \begin{itemize}
     \item baseline phonetic error and hallucination error distribution before and after perturbation are similar 
     \item UU: phonetic error distribution before and after perturbation is similar. The hallucination distribution increases after the perturbation, indicating the hallucination susceptibility of this model. After calculating cosine similarity between the outputs and TF-IDF representation of the training dataset, we discovered that the model does not copy labels from the training set but generates hallucinations unrelated to the content of the training dataset
     \item UR: most of the errors are hallucinations that are concentrated at 0.0 cosine similarity both before and after perturbations, meaning that the sentences are completely semantically disconnected due to repeating labels. Similar phonetic error distribution before and after perturbation
     \item RU: phonetic error distribution before and after perturbation are similar 
     \item RR: different error distribution before and after perturbation. Most of the errors after perturbation have high cosine similarity even though high WER which indicates repetitions
     \end{itemize}

Finally, after evaluating the perturbation method on the CommonVoice dataset, we discover that although it produces hallucinations as expected due to dataset label noise. The perturbation methods tuned for LibriSpeech dataset are not strong enough to induce hallucinations in this dataset, due to the fact that the recording conditions in crowdsourced CommonVoice dataset are very noisy to begin with. The noise levels of the perturbations need to be tuned to the respective test set.

\subsection{Analysis of Source of Hallucination}

We analysed the training dataset to find whether some of the hallucinations are the result of overly memorized phrases. To achieve that, we use TF-IDF statistic for creating representations of transcribed sentences in the dataset and compare it using similarity metric, in our case cosine similarity. In each case, we compare the 5 closest candidates. We discover that most of the hallucinations from UR model are in fact repeated labels. But other models, including highly hallucinatory unique-unique, return low similarity scores, moreover after individual assessment they only sometimes copy certain specific names or terms, that did not appear in the reference sentence but are appearing across the training set.

\section{Conclusion}

Hallucinations are the errors generated by neural ASR models. There are considered to be dangerous because of their seemingly fluid and coherent manner, but lack of connection with the source. We propose a method to distinguish hallucination from phonetic ASR errors using combination of metrics where we implement an algorithm for hallucinatory model classification without the use of training data. Our method allows detecting hallucinatory models in test time by estimating numbers of hallucination before and after noise injection to the utterance. Our method identifies the model that is most susceptible to hallucinations yet has similar WER to baseline. We further explore error distributions and  correlation between hallucination and random label mismatch, and oscillations and the occurrence of repetitive labels.

\section{Limitations}
This work researches English ASR, and although the experiments took into account generalisation across two different datasets, more work regarding generalisation across different languages and speech oriented tasks such as SLU should be considered in the future~\citep{hupkes2023stateoftheart}. Finally, the methods could be re-examined in NLP tasks. 
The methods were used in supervised learning scenarios but could be possibly extended to explore hallucinatory capability of the unsupervised models, i.e. altering the algorithm for hallucination detection in unsupervised models, such as wav2vec2~\citep{BaevskiZMA20}.

The thresholds used for identifying hallucinations and phonetic errors are based on distributions of the errors from the datasets, the heuristic approach that should be automated for better generalisation capability of our methods. Furthermore, our method uses previously extracted filterbank features for the experiments to save preprocessing time during experiments, which limits our work to experimental setting. The online processing is therefore an important future improvement of our method.
Regarding general understanding of hallucinations, the limitation of this approach is combining different automatic metrics together. Creation of a unified metric for measuring  hallucination, especially without the access to the training data, is an important research direction.

In our work we discuss hallucination detection and identification, however we do not concentrate on hallucination mitigation, which is also an important future direction in ASR hallucination research.

\bibliography{custom}

\appendix

\section{ Appendix}
\label{sec:appendix}
\subsection{Model Hyper-parameters}\label{sec:appendix_A1}
 Fairseq ASR transformer consists of 12 encoder layers, 6 decoder layers, with convolutional subsampler and multihead attention with 4 attention heads. The waveform is represented with filterbank features with 80 filters. Models are trained on 2 GPUs GeForce RTX 2080 Ti with 10.761 GB memory. Model hyperparameters learning rate 2e-3, inverse square root for learning rate scheduler, loss function: label smoothed cross entropy with label smoothing 0.1, warm up of 10000 updates and gradient clipping 10.0, maximum tokens 40000 and maximum update value of 300000. We used sentencepiece with unigram vocabulary as bpe tokenizer. We used specaugment with frequency mask F 27, freq mask variable N 2, time mask N of 2, time mask T 100, time mask probability of 1.0 and no time warping. We also used cepstral mean and variance normalization. We average 10 last epochs for testing.
\subsection{Ratio of detected hallucinations in LibriSpeech}\label{sec:appendix_A2}
Ratio of detected hallucinations per model per evaluation dataset. The noisy subsets generate more hallucinatory content, hence the intuition of perturbation as a way to increase probability of hallucinatory output.
\begin{figure}[h!]
  \centering
  \includegraphics[width=1\columnwidth]{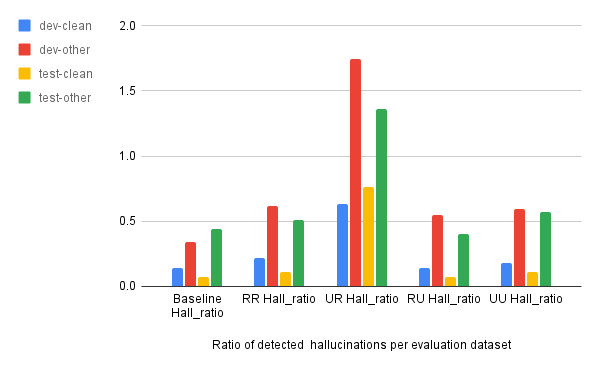}
  \caption{Ratio of detected hallucinations per model per evaluation dataset.   }
\end{figure}

\subsection{Comparison between WER, CHRF2 and BLEU scores}\label{sec:appendix_A3}
Models do not show influence of types of errors on their performance in WER, CHRF2 and BLEU scores apart from the UR model which produced hallucinatory output due to copying labels from the training set.
\begin{table}[h]
\centering
\caption{Comparison between WER, BLEU and CHRF2 Scores of baseline and 4 models trained on 8\% noisy data.}
\label{tab:metric_comparison}
\resizebox{\columnwidth}{!}{%
\begin{tabular}{@{}llll@{}}
\toprule
Libri 360 & WER & BLEU & CHRF2 \\ \midrule
BASELINE & 11.375 & 80.6925 & 90.7225 \\
RR & 15.4075 & 75.6875 & 87.9375 \\
RU & 12.1975 & 78.45 & 89.53 \\
UU & 13.88 & 75.93 & 87.94 \\
UR & 27.405 & 68.6925 & 81.715 \\ \bottomrule
\end{tabular}%
}
\end{table}

\subsection{Impact of Noise Injection Throughout the Sentence on the Model Performance }\label{sec:appendix_A4}
Noise injection in the whole sentence degrades the performance of all the models.

\begin{figure}[h]

  \centering
  \includegraphics[width=1\columnwidth]{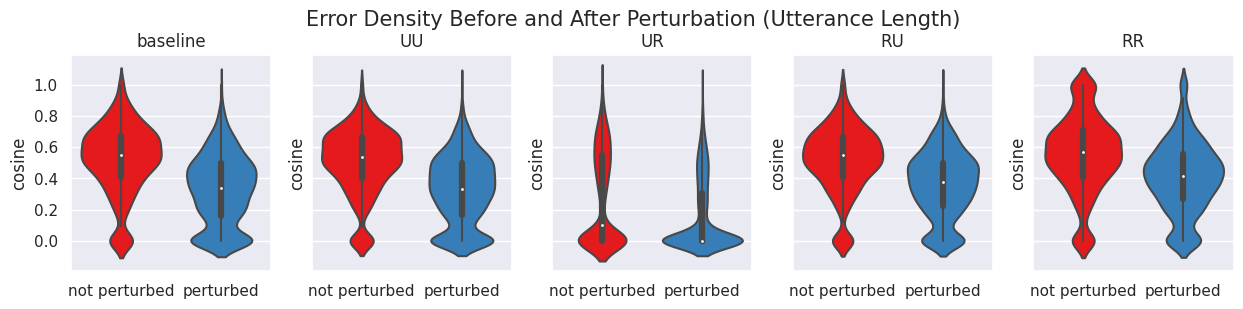}
  \caption{Figure shows decreased cosine similarity of the outputs after perturbation with noise injected throughout the sentence.}

\end{figure}

\end{document}